\documentclass[12pt]{article}
\usepackage{natbib}
\newcommand{\keywords}[1]{\textbf{Keywords:} #1}
\usepackage[ruled,linesnumbered]{algorithm2e}
\usepackage{subcaption}
\usepackage{graphicx,color}
\usepackage{amsmath}
\usepackage{amsthm}
\usepackage{wasysym}
\usepackage{url}
\usepackage{amssymb}
\usepackage{multirow}
\usepackage{bm}
\usepackage{booktabs}
\usepackage{mathrsfs}
\usepackage{fullpage}
\usepackage{setspace}
\usepackage[english]{babel}
\usepackage[utf8]{inputenc}
\usepackage{booktabs}
\usepackage{diagbox}
\usepackage{multirow}
\newtheorem{assumption}{Assumption}

\newtheorem{theorem}{Theorem}
\newtheorem{proposition}{Proposition}

\begin{document}
	
	\title {Community Detection in General Hypergraph via Graph Embedding}
	\author{
			Yaoming Zhen and Junhui Wang\\
	      	School of Data Science\\
			City University of Hong Kong
	       }
	\date{ }
	
	\maketitle
	\onehalfspacing
	
	\begin{abstract}
		 Conventional network data has largely focused on pairwise interactions between two entities, yet multi-way interactions among multiple entities have been frequently observed in real-life hypergraph networks. In this article, we propose a novel method for detecting community structure in general hypergraph networks, uniform or non-uniform. The proposed method introduces a null vertex to augment a non-uniform hypergraph into a uniform multi-hypergraph, and then embeds the multi-hypergraph in a low-dimensional vector space such that vertices within the same community are close to each other. The resultant optimization task can be efficiently tackled by an alternative updating scheme. The asymptotic consistencies of the proposed method are established in terms of  both community detection and hypergraph estimation, which are also supported by numerical experiments on some synthetic and real-life hypergraph networks.
	\end{abstract}
	
	\keywords{Latent space model, network embedding, non-uniform hypergraph, sparse network, tensor decomposition}
	
	\doublespacing
	
	\section{Introduction}
	
	Community detection has attracted increasing attention from both academia and industry in the past few decades, and has a wide spectrum of applications in domains, ranging from social science \citep{ji2016coauthorship, newman2002random, zhao2011community, lee2017time}, life science \citep{chen2006detecting, nepusz2012detecting} to computer science \citep{tron2007benchmark, agarwal2005beyond}. The community structure has been widely observed in many real-life networks, which usually means that entities within the same community tend to interact much more often than across communities. In literature, most conventional network analysis focus on pairwise interactions between two vertices \citep{zhao2012consistency, lei2015consistency, loyal2020statistical, sengupta2018block}. However, the complexity of real-life networks is usually beyond pairwise interaction, and multi-way interactions among vertices arise naturally. For example, in an academic collaboration network, it is often the case that several researches work together on a research project; in a protein-to-protein interaction network, a metabolic reaction usually involves multiple proteins. Under such circumstances, a hypergraph network provides a more faithful representation and retains richer information than a merely vanilla graph network, such as each research project being represented by a hyperedge consisting of multiple vertices being researchers, or each metabolic reaction being a hyperedge consisting of multiple vertices being proteins. In this paper, we are interested in detecting community structure in a hypergraph network, where vertices within the same community share more similar connection patterns compared with vertices in different communities.
	
	To detect communities in a hypergraph network, most existing methods convert the hypergraph into a weighted graph, and then existing graph community detection methods can be applied. For instances, \citet{kumar2018hypergraph} defined the hypergraph modularity as the modularity of the weighted graph and applied standard modularity maximization algorithms for community detection, \citet{lee2020robust} extended the graph-likelihood-based convex relaxation methods \citep{li2018convex} on the adjacency matrix of the weighted graph, and \citet{Ghoshdastidar_2017} conducts spectral clustering on the weighted graph Laplacian for  hypergraph community detection. As pointed out in \citet{dongxia2019}, such conversion will suffer from information loss and lead to suboptimal community detection performance. To circumvent this disadvantage, spiked tensor model \citep{kim2017community} and Tensor-SCORE method \citep{dongxia2019} were proposed to conduct community detection on the hypergraph adjacency tensor directly.
	
	Note that most existing hypergraph community detection methods focus on uniform hypergraph only \citep{GandD2004NIPS, GandD2015AAAI, lee2020robust}, where all the hyperedges consist of exactly the same number of vertices. When it comes to non-uniform hypergraph, they have to decompose a non-uniform hypergraph into a collection of uniform hypergraphs with different orders, due to the difficulty of  lacking an appropriate adjacency tensor for non-uniform hypergraph. It was only until recently that \citet{ouvrard2017adjacency} proposed a heuristic method to construct an adjacency tensor for general hypergraph by adding $m-m_0$ null vertices, where $m$ and $m_0$ are the maximum and minimum cardinality of a hyperedge in the hypergraph, respectively.
	
	In this paper, we propose a novel method for detecting community structure in general hypergraph networks, where a general hypergraph can be uniform or non-uniform. The proposed method consists of an augmentation step and an embedding step. The augmentation step adds a null vertex to the hyperedges with smaller cardinality, and converts the hypergraph into a uniform multi-hypergraph \citep{Pearson2014OnSH, KellyJ.Pearson2015}, which allows vertices to appear multiple times in one hyperedge. The embedding step is formulated in a regularization form with tensor decomposition, which represents each vertex as a low-dimensional numerical vector and encourages vertices within the same community to be close in the embedding space. The proposed hypergraph embedding model is flexible and general, which includes the hypergraph stochastic block model (hSBM), also known as hypergraph planted partition model \citep{Ghoshdastidar_2017}, as its special case. It also accommodates heterogeneity among vertices by allowing vertices within a community to fluctuate in all directions. This is in sharp contrast to hSBM, which assumes vertices within a community share the same spectral embedding vector. It is also more general than the hypergraph degree-corrected block model (hDCBM; \citealt{dongxia2019, yuan2018testing}), under which spectral embeddings of vertices within a community are along the same direction and only differ by their magnitudes. The advantage of the proposed method is supported in both asymptotic theories and numerical experiments on a number of synthetic and real hypergraph networks.
	
	The main contribution of this paper is three-fold. First, we propose a novel hypergraph embedding model (HEM), which consists of an augmentation step and an embedding step. The augmentation step introduces only one null vertex rather than multiple ones as suggested in \citet{ouvrard2017adjacency}, which is more efficient and greatly facilitates the subsequent community detection. The embedding step, to the best of our knowledge, is the first statistical framework that extends the latent space model for graph network to general hypergraph network with theoretical guarantees. Second, a joint modeling framework is developed for simultaneously conducting hypergraph estimation and community detection. Third, we establish the asymptotic consistencies of the proposed method in terms of both hypergraph estimation and community detection in sparse hypergraph network. Particularly, the consistencies hold as long as the link probability is of the order $s_n\gg n^{1-m} \log n$ in an $m$th-order hypergraph with $n$ vertices. This result compares favorably with the existing sparsity results in literature \citep{dongxia2019, Ghoshdastidar_2017, JMLR:v18:16-100}, not to mention that the proposed method also achieves fast convergence rate in  hypergraph estimation.

	The rest of the paper is organized as follows. Section 2 provides a quick review of some preliminaries on hypergraph and tensor. Section 3 presents the details of the proposed hypergraph augmentation and embedding formulations, and an efficient optimization algorithm. Section 4 establishes the asymptotic consistencies of the proposed method. Section 5 examines the numerical performance of the proposed method on both synthetic and real-life hypergraph networks. Section 6 concludes the paper,  and technical proofs and necessary lemmas are contained in the supplementary files.
	
	\section{Preliminaries}
	
\begin{figure}[!htbp]
		\includegraphics[width=8cm,height=6cm]{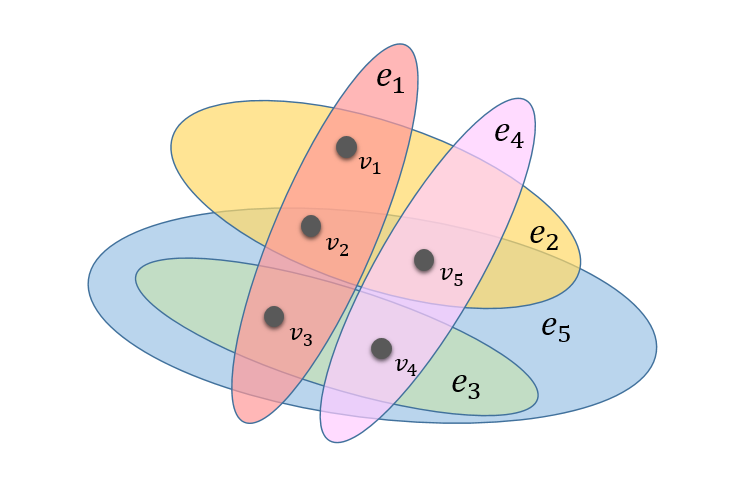}
		\includegraphics[width=8cm,height=6cm]{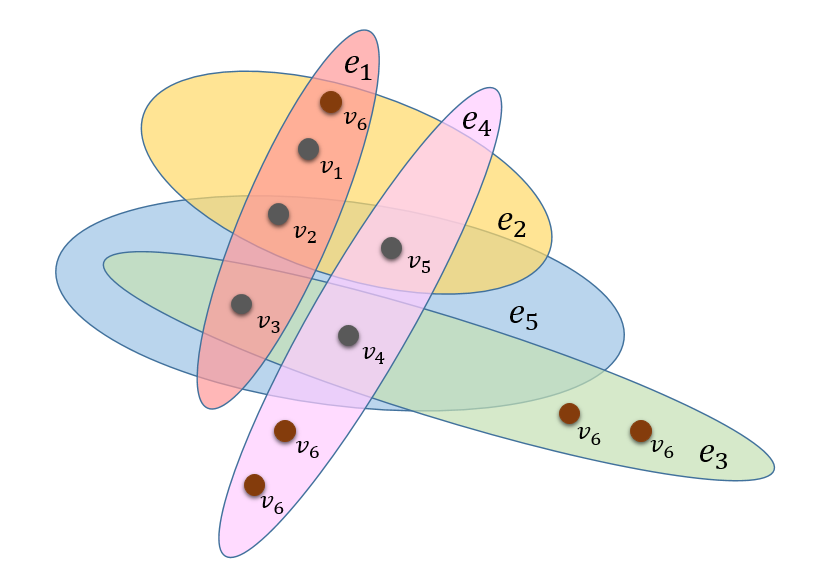}
		\caption{A non-uniform hypergraph with 5 vertices and 5 hyperedges (left) and the corresponding augmented 4-uniform multi-hypergraph (right).}
		\label{fig:illus}
	\end{figure}

	To begin with, we introduce some basic concepts of hypergraph and tensor that will be used extensively in the sequel. A general hypergraph is denoted as $\mathcal{H}(V,E)$, where $V=\{v_1,v_2,...,v_n\}$ is a vertex set with $n$ vertices, $E$ consists of all hyperedges, and each hyperedge may contain multiple vertices in $V$. Denote $[n]=\{1,2,...,n\}$, and we set $V=[n]$ for simplicity, and thus a hyperedge is a non-empty subset of $[n]$. A hypergraph is called $m$-uniform if the cardinality of every hyperedge equals $m$. On the flip side, a non-uniform hypergraph contains hyperedges whose cardinalities can vary from one to another. A simple non-uniform hypergraph with 5 vertices and 5 hyperedges is displayed in the left panel of Figure \ref{fig:illus}, where hyperedges $e_1$ and $e_2$ consist of 3 vertices, hyperedges $e_3$ and $e_4$ consist of 2 vertices, and hyperedge $e_5$ consists of 4 vertices. Note that the hypergraphs considered in this paper are undirected; that is, no ordering of vertices is necessary within the hyperedges. For definition of directed hypergraph, interested readers may refer to \cite{directedhg1993} and \cite{ouvrard2017adjacency}.
	
	
	A tensor $\mathcal{A}=(a_{i_1...i_m})\in \mathbb{R}^{I_1\times ...\times I_m}$  is a cubical tensor if $I_1=...=I_m=n,$ for some $n\in \mathbb{Z}^+$. More formally, such a tensor is called an $m$th-order tensor of dimension $n$. A cubical tensor $\mathcal{A}$ is super symmetric if $a_{i_1...i_m}=a_{\pi(i_1)...\pi(i_m)}$, for any possible permutations $\pi \in S_m$, the symmetric group of degree $m$. A $m$th-order super symmetric tensor $\mathcal{I} \in \{0,1\}^{n\times...\times n}$ is called an identity tensor if $I_{i_1...i_m}=1$ when $i_1=...=i_m$ and 0 otherwise.
	
	Let $M^{(j)}\in \mathbb{R}^{Q_j\times I_j}$ be $m$ matrices, then the mode $j$ product \citep{SIAMreview} between $\mathcal{A}$ and $M^{(j)}$ is defined as $\mathcal{A}\times_j M^{(j)}\in \mathbb{R}^{I_1\times...\times I_{j-1}\times Q_j\times I_{j+1}...\times I_m}$ with entries
	\begin{equation*}
		(\mathcal{A}\times_j M^{(j)})_{i_1...i_{j-1}q_j i_{j+1}...i_m}=\sum\limits_{i_j=1}^{I_j}a_{i_1...i_{j-1}i_ji_{j+1}...i_m}M^{(j)}_{q_ji_j}.
	\end{equation*}
	Moreover, $\mathcal{A}\times_1M^{(1)}\times_2...\times_mM^{(m)}\in \mathbb{R}^{Q_1\times...\times Q_m}$ is such a tensor with entries
	\begin{equation*}
		(\mathcal{A}\times_1M^{(1)}\times_2...\times_mM^{(m)})_{q_1...q_m}=\sum\limits_{i_1=1}^{I_1}...\sum\limits_{i_m=1}^{I_m}a_{i_1...i_m}M^{(1)}_{q_1i_1}...M^{(m)}_{q_mi_m}.
	\end{equation*}
	
	Recall that a multi-set $\{\{i_1,i_2,...,i_m\}\}$ is an extension of set in that some elements may appear multiple times. Herein,  to distinguish from a set, we use double curly braces $\{\{\cdot\}\}$ to denote a multi-set as in \citet{bahmanian2015connection} and \citet{kovavcevic2018codes}. Clearly, if the elements $i_1, ..., i_m$ are distinct with one another, the multi-set $\{\{i_1, ..., i_m\}\}$ reduces to the set $\{i_1, .., i_m\}$. For any $i_1,...,i_m\in [n+1]$, we define an augmented Kronecker delta to be $\delta^{n+1}_{i_1...i_m}=0$ if $\{\{i_1,...,i_m\}\}\backslash\{n+1\}$ is indeed a non-empty set and 1 otherwise, where the multi-set difference with a set $\{\{i_1, ..., i_m\}\}\setminus \{n+1\}$ is a multi-set containing elements in the multi-set $\{\{i_1, ..., i_m\}\}$ but not in the set $\{n+1\}$. For example, if $n=100$ and $m=3$, we have $\delta^{101}_{1, 2, 101} = 0$, $\delta^{101}_{101, 101, 101} = 1$ and $\delta^{101}_{1, 1, 2} =1$. Furthermore, the order-specified augmented Kronecker delta $\delta^{n+1,ord}_{i_1...i_m}=0$ if and only if $\delta^{n+1}_{i_1...i_m}=0$ and $i_1\le ...\le i_m$. These two extended Kronecker deltas can greatly simplify notations and formulations for non-uniform hypergraph, and they are slightly different from the vanilla extension in \citet{QI20051302}, where the extended Kronecker delta equals 0 if there exist at least 2 distinct elements.
	
	\section{Community detection in hypergraph}
	
	This section proposes a novel community detection method for general hypergraphs. The key idea is to introduce a null vertex  so that hyperedges with smaller cardinality can be augmented, and the non-uniform hypergraph is transformed into a special uniform multi-hypergraph, where the null vertex may appear multiple times in some hyperedges. Then the proposed method proceeds to seek for a suitable numeric embedding that embeds all vertices of the hypergraph into a low dimensional Euclidean space, where vertices belong to the same community tend to have shorter distance.
	
	
	\subsection{Hypergraph augmentation}
	
	Let $\mathcal{H}(V,E)$ be a hypergraph, possibly non-uniform, with $V=[n]$ and range $m\ge2$, which is the maximum cardinality of a hyperedge in $E$. One of the key challenges of analyzing non-uniform hypergraph is due to the unequal hyperedge cardinalities, leading to the ambiguity of discriminating hyperedges and their proper subsets. 
	
	To circumvent this difficulty, we introduce a null vertex, denoted as $v_{n+1}$, and then any hyperedge with cardinality less than $m$ can be augmented to a multi-set with $m$ elements. For instance, when $m>3$, a hyperedge with vertices $\{1,2,3\}$ can be augmented to $\{\{1,2,3,n+1,\ldots,n+1\}\}$ with $m-3$ null vertices, whereas another hyperedge with vertices $\{1,\ldots,m\}$ stays the same without any additional null vertex. It is clear that $\cal H$ is converted to an equivalent $m$-uniform multi-hypergraph, where each hyperedge is a multi-set with cardinality $m$ and only the null vertex $v_{n+1}$ is allowed to appear multiple times in a hyperedge. With slight abuse of notation, we still denote this $m$-uniform multi-hypergraph as $\cal H$. A simple illustration of this augmentation step is displayed in the right panel of Figure \ref{fig:illus}.
	
	We remark that when $\cal H$ is a uniform hypergraph, the augmentation step and the subsequent special treatments of the null vertex are not necessary. For ease of presentation, we assume $\cal H$ is a non-uniform hypergraph hereafter, and the upcoming proposed model can be adapted to uniform hypergraph with slight modification. As such, we use the $m$th-order adjacency tensor  $\mathcal{A}=(a_{i_1...i_m})\in \{0,1\}^{(n+1)\times...\times (n+1)}$ to represent $\mathcal{H}$ with entries
	\begin{equation*}
		a_{i_1...i_m}=
		\begin{cases}
			1 & \text{if } \{\{i_1,...,i_m\}\}\setminus\{n+1\}\in E;\\
			0 & \text{otherwise.}
		\end{cases}
	\end{equation*}
	Apparently, a necessary condition for $a_{i_1...i_m}=1$ is $\delta_{i_1...i_m}^{n+1}=0$. It is clear that this augmentation step, converting a non-uniform hypergraph to a uniform multi-hypergraph, is critical to make downstream analyses more statistically and computationally tractable.
	
	Suppose there are $K$ potential communities within the hypergraph, and let $\psi:[n]\longrightarrow [K]$ be the community assignment function, then $\psi(i)=s$ if vertex $v_i$ belongs to the $s$-th community. We also write $\psi_i=\psi(i)$ for short. Note that the null vertex $v_{n+1}$ does not belong to any community, and thus we set $\psi_{n+1}=K+1$ for formality. Denote $Z=(z_{ij})\in \mathbb{R}^{(n+1)\times (K+1)}$ to be the corresponding community membership matrix with $z_{ij}=1$ if $j=\psi_i$ and 0 otherwise. It is obvious that $z_{i,K+1}=1$ only when $i=n+1$.
	
	
	\subsection{Hypergraph embedding}
	
	Given the $m$-uniform multi-hypergraph $\cal H$ from the augmentation step, we let $\mathcal{P}=(p_{i_1...i_m})\in (0,1)^{(n+1)\times...\times (n+1)}$ be the $m$th-order probability tensor, with $p_{i_1...i_m}=P(a_{i_1...i_m}=1)$ if $\delta^{n+1}_{i_1...i_m}=0$, and let $\Theta=(\theta_{i_1...i_m})\in\mathbb{R}^{(n+1)\times...\times (n+1)}$ be the entrywise transformation of $\mathcal{P}$ so that
	\begin{equation}
		\label{logit}
		\theta_{i_1...i_m}=\log \Big ( \frac{p_{i_1...i_m}}{s_n-p_{i_1...i_m}} \Big ),
	\end{equation}
	where the sparsity factor $s_n$ that may vanish with $n$ is introduced to accommodate sparse networks. The modified logit transformation \eqref{logit} implies that $p_{i_1 \ldots i_m} = s_n(1+e^{-\theta_{i_1 \ldots i_m}})^{-1}$, and the number of hyperedges is of order $O_p(s_nn^m)$ if $\Theta$ lies in a compact subset of $\mathbb{R}^{(n+1)\times...\times (n+1)}$.  We remark that many frequentist analysis of network models multiplies the network underlying linking probability by a decaying sparsity coefficient to accommodate sparse networks, such as the hypergraph stochastic block model \citep{Ghoshdastidar_2017, JMLR:v18:16-100} and hypergraph degree-corrected block model \citep{dongxia2019}. Yet this treatment may not be directly applied to latent space model with standard logit transformation, since tiny value of $p_{i_1...i_m}$ implies that $\theta_{i_1...i_m}$ will be pushed towards $-\infty$, and thus the entries of $\bm{\alpha}$ may diverge to $\pm \infty$, leading to unstable numerical performance in estimating $\bm{\alpha}$. 
	
	
	We now turn to embed the hypergraph into an $r$-dimensional Euclidean space with $2\le r\ll n$, where each vertex is represented by an $r$-dimensional vector $\bm{\alpha}_i$ for any $i \in [n]$. The embedding dimension $r$ here is allowed to diverge with $n$. For the null vertex, we simply set $\bm{\alpha}_{n+1}=r^{-1/2}\bm{1}_r$, the $r$ dimensional vector with every entry $r^{-1/2}$. Let $\bm{\alpha} = (\bm{\alpha}_1,...,\bm{\alpha}_{n+1})^T$ be the embedding matrix. We consider the following hypergraph embedding model (HEM),
	\begin{equation}
		\Theta=\mathcal{I}\times_1\bm{\alpha}\times_2...\times_m\bm{\alpha},
		\label{LFM_Tensor}
	\end{equation}
	where $\mathcal{I}\in \mathbb{R}^{r\times...\times r}$ is the $m$th-order identity tensor of dimension $r$. Clearly, for any $i_1,...,i_m \in [n+1]$, model (\ref{LFM_Tensor}) assumes that the information contained in $\theta_{i_1...i_m}$ can be fully captured by the embedding matrix $\bm{\alpha}$ in that
	$\theta_{i_1...i_m}=\mathcal{I}\times_1\bm{\alpha}^T_{i_1}\times_2...\times_m\bm{\alpha}^T_{i_m}$. When $\delta_{i_1...i_m}^{n+1, ord}=0$, the probability that $\{\{i_1, ..., i_m\}\}\setminus \{n+1\}$ forms a hyperedge is $s_n(1+e^{-\theta_{i_1...i_m}})^{-1}$, whereas $\theta_{i_1...i_m}$ or $p_{i_1...i_m}$ defined in (\ref{LFM_Tensor}) does not have any probability interpretation when $\delta_{i_1...i_m}^{n+1, ord}= 1$, and is thus inconsequential.
	
	The proposed HEM is flexible and general, and includes the celebrated hSBM \citep{Ghoshdastidar_2017} as its special case. Particularly, taking $\bm{\alpha} = ZC$, HEM reduces to $\Theta = \mathcal{B}\times_1 Z\times_2...\times_m Z$ with $\mathcal{B} = \mathcal{I}\times_1 C \times_2...\times_m C$, which becomes an hSBM with membership matrix $Z\in \{0, 1\}^{(n+1)\times (K+1)}$ and a transformed core probability tensor $\mathcal{B}\in \mathbb{R}^{(K+1)\times...\times (K+1)}$. Yet, HEM is more flexible as it naturally accommodates heterogeneity among vertices by allowing the embeddings or spectral embeddings of vertex within a community to fluctuate in all directions. It is worth pointing out that such accommodation of heterogeneity is more general than that in hDCBM \citep{dongxia2019, yuan2018testing}, which requires spectral embeddings of vertices within a community to lie along the same direction and only differ by magnitudes. 
	
	
	It is also interesting to remark that HEM is equivalent to assuming that the super symmetry tensor $\Theta$ has a symmetric CP decomposition \citep{SIAMreview}, $\Theta=\sum\limits_{j=1}^r\bm{\alpha}_{\cdot j}\circ...\circ \bm{\alpha}_{\cdot j}$, where $\alpha_{\cdot j}$ is the $j$-th column of $\bm{\alpha}$ and $\circ$ stands for the vector outer product. Furthermore, we define the symmetric rank of an $m$th-order tensor $\Theta$ of dimension $n+1$ over $\mathbb{R}$ as ${\rm rank}(\Theta) = \min \{R: \Theta = \sum_{r=1}^{R}u_r\circ...\circ u_r, u_r\in \mathbb{R}^{n+1} \}$. Therefore, it is assumed that $\Theta$ has symmetric rank at most $r$. We also remark our definition of symmetric rank is slightly different from the one that defined in literature \citep{SIAMreview, Robeva2016} by requiring the embedding vectors to be real.
	
	More importantly, HEM is identifiable if $\Theta$ has symmetric rank $r$ and $m\ge 3$,  as the factorization (\ref{LFM_Tensor}) is unique up to column permutations of $\bm{\alpha}$ \citep{sidiropoulos2000uniqueness}. When $\Theta$ has symmetric rank $r$ but $m = 2$,  HEM reduces to the latent space model for graph networks \citep{hoff2002latent} and the factorization  (\ref{LFM_Tensor}) is unique up to any orthogonal transformation of $\bm{\alpha}$. Note that column permutations or general orthogonal transformations are isometric linear transformation of the rows of $\bm{\alpha}$, the community structure encoded in $\bm{\alpha}$ always remains unchanged.
	
	\subsection{Penalized log-likelihood objective}
	
	With the factorization of $\Theta$ in (\ref{LFM_Tensor}), the negative log-likelihood function of $\mathcal{H}$ becomes
	$$
	\mathcal{L}(\bm{\alpha};\mathcal{A})=\frac{1}{\varphi(n,m)} \sum_{\delta_{i_1...i_m}^{n+1,ord}=0}L(\theta_{i_1...i_m};a_{i_1...i_m}),
	$$
	where $\varphi(n,m)=\sum\limits_{k=1}^m\binom{n}{k}$ is the number of potential hyperedges with $\delta_{i_1...i_m}^{n+1,ord}=0$, $\theta_{i_1...i_m}=\mathcal{I}\times_1\bm{\alpha}^T_{i_1}\times_2...\times_m\bm{\alpha}^T_{i_m}$, and
	\begin{equation*}
		L(\theta_{i_1...i_m};a_{i_1...i_m}) = \log \Big(1+\frac{s_n}{1-s_n+e^{-\theta_{i_1...i_m}}} \Big) - a_{i_1...i_m}\log \Big(\frac{s_n}{1-s_n+e^{-\theta_{i_1...i_m}}} \Big).
	\end{equation*}
	
	We next equip $\mathcal{L}(\bm{\alpha}; \mathcal{A})$ with a novel penalty term to enhance the feasibility of computation and hypergraph community detection. This leads to the proposed regularized cost function,
	\begin{equation}
		\label{rnll} \mathcal{L}_\lambda(\bm{\alpha};\mathcal{A})=\mathcal{L}(\bm{\alpha}; \mathcal{A})+\lambda_nJ(\bm{\alpha}),
	\end{equation}
	where $\lambda_n$ is a positive tuning parameter and
	\begin{equation}
		\label{J}
		J(\bm{\alpha})= \min_{Z\in\Gamma, C\in \mathcal{R}^{(K+1)\times r}}\frac{1}{n}||\bm{\alpha} - ZC||_F^2,\text{ } C_{K+1} = r^{-1/2}\bm{1}_r,
	\end{equation}
	is introduced to encourage the community structure encoded in $\bm{\alpha}$. Herein, $\Gamma$ is the set of all possible community membership matrix. That is, for any $Z \in \Gamma \subset \mathbb{R}^{(n+1)\times (K+1)}$, each row of $Z$ contains exactly one 1 with all other entries being zeros, and $Z_{i(K+1)} = 1$ if and only if $i = n+1$. It is clear that the embeddings of vertices with similar linking patterns will be pushed towards the same center, and thus close to each other in the embedding space, leading to the desired community structure in $\cal H$. A similar regularization term has been employed in \citet{tang2019individualized} for individualized variable selection and in \citet{zhang2021Directed} for community detection in directed networks. In the sequel, with $\Theta = \mathcal{I} \times_1 \bm{\alpha}...\times_m \bm{\alpha}$, we use $\mathcal{L}_\lambda(\Theta; \mathcal{A})$ and $\mathcal{L}_\lambda(\bm{\alpha}; \mathcal{A})$ interchangeably, as convenience dictates.  
	
	We develop an alternative updating scheme to minimize (\ref{rnll}). Particularly, given $Z^{(t)}$ and $C^{(t)}$ at step $t$, $\bm{\alpha}$ can be updated by solving
	\begin{equation*}
		\underset{\bm{\alpha}_{n+1} = r^{-1/2}\bm{1}_r}{\min}\text{ } \frac{1}{\varphi(n,m)}\underset{\delta^{n+1,ord}_{i_1...i_m}=0}{\sum}L(\theta_{i_1...i_m};a_{i_1...i_m})+\frac{\lambda_n}{n}\|\bm{\alpha}-Z^{(t)}C^{(t)}\|_F^2.
	\end{equation*}
	Denote $\mathcal{T} = \frac{\partial{\mathcal{L}(\Theta, \mathcal{A})}}{\partial\Theta}$, and let $\Delta = \{0, 1\}^{(n+1)\times...\times (n+1)}$ such that $\Delta_{i_1,,,i_m} = 1$ if $\delta_{i_1 i_2...i_m}^{n+1} = 0$ with $\delta_{i_2 i_3...i_m(n+1)}^{n+1, ord} = 0$ or $\delta_{i_1(n+1)...(n+1)}^{n+1} = 0$, and 0 otherwise. We then update the first $n$ rows of $\bm{\alpha}$, which is denoted as $\bm{\alpha}_{1:n}$, along its gradient, $\bm{\alpha}_{1:n}^{(t+1)}=\bm{\alpha}_{1:n}^{(t)}-\eta_t\nabla_{\bm{\alpha}_{1:n}}\mathcal{L}_\lambda^{(t)}(\bm{\alpha}^{(t)})$, where $\eta_t>0$ is the learning rate at step $t+1$, and
	$$
	\nabla_{\bm{\alpha}_{1:n}}\mathcal{L}_\lambda^{(t)}(\bm{\alpha}^{(t)})=\frac{1}{\varphi(n,m)}\langle \mathcal{T}*\Delta, \mathcal{I}\times_2 \bm{\alpha} \times_3...\times_m \bm{\alpha}\rangle^{\{2, ..., m\}}_{1:n} + \frac{2\lambda_n}{n}(\bm{\alpha}-Z^{(t)}C^{(t)})_{1:n}.
	$$
	Herein, $*$ is the Hadamard product (entry-wise product) between two tensors, and $\langle \mathcal{T}*\Delta, \mathcal{I}\times_2 \bm{\alpha} \times_3...\times_m \bm{\alpha}\rangle^{\{2, ..., m\}}\in \mathbb{R}^{(n+1)\times r}$ is the tensor inner product between $\mathcal{T} * \Delta$ and $ \mathcal{I}\times_2 \bm{\alpha} \times_3...\times_m \bm{\alpha}$ with respect to the second, ..., $m$-th modes. Specifically, the $(i, j)$-th entry of $\langle \mathcal{T}*\Delta, \mathcal{I}\times_2 \bm{\alpha} \times_3...\times_m \bm{\alpha}\rangle^{\{2, ..., m\}}$ is $\sum_{i_2, ..., i_m} (\mathcal{T}*\Delta)_{ii_2...i_m}(\mathcal{I}\times_2 \bm{\alpha}\times_3...\times_m \bm{\alpha})_{ji_2...i_m}$.
	
	Next, given $\bm{\alpha}^{(t+1)}$, the sub-optimization task now becomes
	\begin{equation}
		\label{center_update}
		\min_{Z\in\Gamma, C\in \mathcal{R}^{(K+1)\times r}}\frac{1}{n}||\bm{\alpha}^{(t+1)} - ZC||_F^2, \text{ subject to } C_{K+1} = r^{-1/2}\bm{1}_r.
	\end{equation}
	Clearly, it resembles the K-means formulation for $\bm{\alpha}^{(t+1)}_{1:n}$, and thus a standard K-means algorithm can be employed to solve for $Z$ and $C$.
	
	As computational remarks, the alternative updating algorithm is guaranteed to converge to a stationary point, and its computational complexity is of order $O\big(\kappa_2 (n^mr + \kappa_1 K nr)\big)$, where $\kappa_1$ is the number of iterations for the $K$-means algorithm and  $\kappa_2$ is the number of iterations for the gradient descent step. Note that low rank tensor approximation tends to be computationally expensive and easy to get trapped in non-informative local minima \citep{Arous2019}. Based on our limited numerical experience, a warm initialization of $\bm{\alpha}^{(0)}$ can greatly help with the numerical convergence. In all the numerical examples, we initialize $\bm{\alpha}^{(0)}$ with a higher-order singular value decomposition algorithm (HOSVD; \citealp{HOSVD2000}). We also suggest to set the embedding dimension $r = K$ in practice, similar suggestions were also made for some spectral-clustering-based algorithms \citep{Ghoshdastidar_2017}. When $K$ is unknown, we can follow the procedure in \citet{dongxia2019} to investigate the ``eigen-gap" of the network adjacency tensor. Specifically, one can first obtain a spectral embedding $\widehat{U}$ with sufficiently large dimension, and then investigate the eigen-gap of the matrix $\mathcal{M}_1(\mathcal{A} \times_3 \widehat{U} \times_4 ... \times_m \widehat{U})$. Other data adaptive selection criteria, such as network cross-validation  \citep{doi:10.1080/01621459.2016.1246365, zhuji2020} may be employed as well, at the cost of increased computational burden.  
	
	\section{Asymptotic theory}
	
	This section establishes some theoretic results to quantify the asymptotic behavior of the proposed HEM method in  estimating the underlying transformed probability tensor $\Theta$ as well as detecting community structure in a general hypergraph.

	\subsection{Consistency in estimating $\Theta^*$}
	
	Let $\Omega = \{\Theta : \Theta = \mathcal{I}\times_1\bm{\alpha} \times_2 ...\times_m \bm{\alpha}, \max_{i\in [n+1]}\|\bm{\alpha}_i \|_2\le c_0, \bm{\alpha}_{n+1} = r^{-1/2}\bm{1}_r\}\subset \mathbb{R}^{(n+1)\times ... \times (n+1)}$ be the domain of the problem, for a positive constant $c_0\ge 1$. It is clear that $\Omega$ is a compact subset of $\mathbb{R}^{(n+1)\times...\times (n+1)}$, and for any $\Theta \in \Omega$, it has symmetric rank at most $r$. Denote $\bm{\alpha}^*$ as the underlying true hypergraph embedding with $\Theta^* = \mathcal{I} \times_1 \bm{\alpha}^*\times_2...\times_m \bm{\alpha^*} \in \Omega$. In addition, for any $\Theta \in \Omega$, we define $e_L(\Theta,  \Theta^*) = \varphi^{-1}(n,m)\sum_{\delta_{i_1...i_m}^{n+1,ord}=0} E\big(L(\theta_{i_1...i_m};a_{i_1..i_m}) - L(\theta^*_{i_1...i_m};a_{i_1...i_m})\big)$, which is the average of certain Kullback-Leibler divergences and guaranteed to be non-negative.
	
	The following large deviation inequality is derived to quantify the asymptotic behavior of $\mathcal{L}_\lambda(\Theta;\mathcal{A})$ in the neighborhood of $\Theta^*$ defined by $e_L(\Theta, \Theta^*)$.
	
	\begin{proposition}
		\label{proposition1}
		Suppose $\lambda_n J(\bm{\alpha}^*)< \frac{1}{2}\epsilon_n$, then there exists some absolute constants $c_1, c_2$ such that if $nr\varphi^{-1}(n,m) \epsilon_n^{-1} \log \epsilon_n^{-1/2} \le c_1$, we have
		\begin{equation*}
			P\Big(\sup_{\{\Theta \in \Omega| e_L(\Theta,\Theta^*) \ge \epsilon_n\}}\big(\mathcal{L}_\lambda(\Theta^*;\mathcal{A})-\mathcal{L}_\lambda(\Theta;\mathcal{A})\big)\ge 0 \Big) \le 2 \exp\big(-c_2\varphi(n,m) \epsilon_n\big).
		\end{equation*}
	\end{proposition}
	
 Proposition \ref{proposition1} gives an intermediate result for establishing estimation consistency in Theorem \ref{theorem1}, and it assures that any $\Theta$ such that $\mathcal{L}_\lambda(\Theta;\mathcal{A}) \le \mathcal{L}_\lambda(\Theta^*;\mathcal{A})$ shall lie in  the neighborhood of $\Theta^*$ with high probability. By the condition that $nr\varphi^{-1}(n,m) \epsilon_n^{-1} \log \epsilon_n^{-1/2} \le c_1$, the fastest order of $\epsilon_n$ can be set as $\epsilon_n  = \frac{nr\log n}{\varphi(n, m)}$ , which is governed by the network size $\varphi(n,m)$ and the number of parameters $nr$ in HEM. It is interesting to remark that $\lambda_n$ appears to have no effect on the order of $\epsilon_n$, as long as it satisfies the condition $\lambda_n J(\bm{\alpha^*})\le \frac{1}{2}\epsilon_n$.
	
We are now ready to establish the consistency of $\widehat{\Theta} = \mathcal{I}\times_1\hat{\bm{\alpha}}\times_2...\times_m \hat{\bm{\alpha}}$, with $\hat{\bm{\alpha}}$ being the estimate from Section 3.3.  Let $p(y;\theta)$ be the density of a Bernoulli random variable with parameter $p=s_n(1+\exp(-\theta))^{-1}$. Then the discrete Hellinger distance between $p(y;\theta)$ and $p(y;\theta^*)$ is defined as
	\begin{equation*}
		d(\theta,\theta^*)= \Big [ \big (p^{1/2}-(p^*)^{1/2}\big )^2+
		\big ((1-p)^{1/2}-(1-p^*)^{1/2}\big )^2 \Big ]^{1/2},
	\end{equation*}
and the deviation between $\Theta$ from $\Theta^*$ can be evaluated by the averaged squared Hellinger distance,
	\begin{equation*}
		D^2(\Theta,\Theta^*)=\frac{1}{\varphi(n,m)}\sum_{\delta^{n+1,ord}_{i_1...i_m}=0}d^2(\theta_{i_1...i_m},\theta^*_{i_1...i_m}).
	\end{equation*}
	
	
	\begin{theorem}
		\label{theorem1}
		Under the assumptions in Proposition 1, for any $\widehat{\Theta}$ with $\mathcal{L}_\lambda (\hat{\bm{\alpha}}; \mathcal{A}) \le \mathcal{L}_\lambda (\bm{\alpha}^*; \mathcal{A})$, we have
		$$
			P\big(D^2(\widehat{\Theta},\Theta^*)\ge \epsilon_n)\le 2 \exp\big(-c_2\varphi(n,m) \epsilon_n\big),
		$$
		Moreover, $D^2(\widehat{\Theta},\Theta^*)=O_p(\epsilon_n)$ and $n^{-m/2}\|\widehat{\Theta}-\Theta^*\|_F=O_p(\sqrt{\epsilon_n/s_n})$.
	\end{theorem}

	Theorem 1 shows that  a reasonably good solution $\widehat{\Theta}$ is guaranteed to converge to $\Theta^*$ at a fast rate, which depends on the centrality of the community structure encoded in $\bm{\alpha}^*$, the network size and number of parameters via $\epsilon_n$, and the network sparsity factor $s_n$. The condition $\mathcal{L}_\lambda (\hat{\bm{\alpha}}; \mathcal{A}) \le \mathcal{L}_\lambda (\bm{\alpha}^*; \mathcal{A})$ shall be satisfied by the solution obtained in Section 3.3, when the estimation algorithm is initialized by some value in a small neighborhood of $\Theta^*$. The consistency result in Theorem 1 holds true with $O(rn^{1-m}\log n) \le \epsilon_n \ll s_n$. If we further set $r < \log n$, this yields a slightly weaker sparsity assumption than that in \cite{dongxia2019} and \cite{Ghoshdastidar_2017, JMLR:v18:16-100}, where the smallest sparsity factor is of the order $n^{1-m} (\log n)^2$ for some fixed $K$. 
	
	
	\subsection{Consistency in community detection}
	
	We now turn to establish the consistency of community detection for general hypergraphs. Let $\psi^*:[n]\longrightarrow [K]$ be the true community assignment function, and $\hat{\psi}$ be the estimation counterpart induced by $\hat{\bm{\alpha}}$. Formally, $\hat{\psi}_i = \arg\min_{k\in [K]}||\bm{\alpha}_i-\widehat{C}_k||_2$, where $\widehat{C} = \arg\min_C\min_{Z\in \Gamma} ||\hat{\bm{\alpha}} - ZC||_F$, subject to $C_{K+1} = r^{-1/2}\bm{1}_r$.  The community detection error of $\hat{\psi}$ can be evaluated by the minimum scaled Hamming distance between $\hat{\psi}$ and $\psi^*$ under permutations, which is defined as
	\begin{equation}
		\label{hamming}
		\text{err}(\psi^*,\hat{\psi})=\min_{\pi\in S_K}\frac{1}{n}\sum\limits_{i=1}^n\bm{1}\{\psi^*_i\ne \pi(\hat{\psi}_i)\},
	\end{equation}
	where $\bm{1}\{\cdot\}$ is the indicator function and $S_K$ is the symmetric group of degree $K$. Clearly, it measures the minimum fraction of vertices that are misclassified by $\hat{\psi}$ under permutation, and $\hat{\psi}$ is a consistent estimator of $\psi^*$ if $ err(\psi^*,\hat{\psi})$ goes to zero with probability tending to 1. Such a scaled or unscaled Hamming distance has become a popular metric in quantifying the performance of community detection \citep{JMLR:v18:16-100, Ghoshdastidar_2017, dongxia2019, jing2020community, lee2020robust}.
	
	
	Let $N^*_k=\{i:\psi^*_i=k\}$ be a true network community with cardinality $n_k$ and $C^*\in \mathbb{R}^{(K+1)\times r}$ be the associated community embedding center matrix in the sense that $C^*_k = \frac{1}{n_k}\sum_{\psi^*_i=k} \bm{\alpha}^*_i$, for $k\in [K]$, and $C_{K+1}^* = r^{-1/2}\bm{1}_r$. Denote $\mathcal{B}^* = \mathcal{I}\times_1 C^* \times_2...\times_m C^*$. The following assumptions are made to ensure that communities within the hypergraph networks are asymptotically identifiable.
	\begin{assumption}
		There exists a constant $c_3>0$ such that
		\begin{equation*}
			\min_{k,k^\prime \in [K],k\ne k^\prime} K^{(1-m)/2} \|\mathcal{B}^*_{k}-\mathcal{B}^*_{k^\prime}\|_F\ge c_3\gamma_n,
		\end{equation*}
		where $\mathcal{B}^*_{k}$ is the $k$-th sub-tensor of $\mathcal{B}^*$ by fixing the first index as $k$, and $\gamma_n$ may converge to 0 with $n$.
	\end{assumption}

	\begin{assumption}
		There exists a constant $c_4$ such that $\max_k n_k \le c_4 \min_k n_k$.
	\end{assumption}
	
	Assumption A is an identifiability assumption, which assumes that the true communities are well separated as $n$ grows, and is crucial to the feasibility of community detection. It is interesting to remark that a signal-to-noise ratio is used in \citet{yuan2018testing} to characterizes the separability among communities under hSBM. A similar community separation assumption can be found in \citet{Leijing2020} for multi-layer network model. Assumption B assures the communities are well defined and will not degenerate asymptotically. This assumption is mild and satisfied when the vertex community memberships come from a multinomial distribution. The same assumption can be found in \citet{dongxia2019}, and a relatively stronger assumption can be found in \citet{chien2019minimax} assuming equal community sizes.
	
	
	\begin{theorem}
		Suppose all the assumptions in Theorem 1 as well as Assumptions A and B are satisfied,
		$\underset{n\rightarrow +\infty}{\lim}\lambda_n \epsilon_n s_n^{-2} (\log s_n^{-1})^{-1}> 0 $ and $K = o(\gamma_n^2s_n\epsilon_n^{-1})$, then $\text{err}(\psi^*, \hat{\psi})=O_p(\epsilon_ns_n^{-1}\gamma_n^{-2})$.
	\end{theorem}
	
	Theorem 2 assures that the community structure in a general hypergraph can be consistently recovered by the proposed HEM method. The consistency result holds true for diverging $K$ as long as it does not diverge too fast. Furthermore, Theorem 2 requires that $\lambda_n$ cannot be too small, whereas Theorem 1 requires $\lambda_n$ to be sufficiently small. Combining these two gives a proper interval for the order of $\lambda_n$ to assure consistency in both network estimation and community detection.
	
As a theoretical example, consider a hypergraph network with $J(\bm{\alpha}^*) \le \epsilon_n^2 \big(s_n^2 \log( s_n^{-1})\big)^{-1}$ and $r$, $\gamma_n$ and $K$ are of the constant order. With $\lambda_n = \frac{1}{2}\epsilon^{-1}_ns_n^2\log (s_n^{-1})$, Theorem 1 implies that $\epsilon_n$ is of the order $n^{1-m} \log n$, and Theorem 2 further implies that $\text{err}(\psi^*, \hat{\psi})= O_p(\log n / (n^{m-1} s_n))$, which matches up with the error rates in \citet{Ghoshdastidar_2017, JMLR:v18:16-100} and \citet{dongxia2019}. To ensure the community detection consistency, Theorem 2 requires that $s_n \gg \log n/ n^{m-1}$, which is slightly weaker than the sparsity requirement in \citet{Ghoshdastidar_2017, JMLR:v18:16-100} and \citet{dongxia2019}. We also remark that under homogeneous hSBM, where the probability of any $m$ vertices forming a hypergraph is $p$ if they are from the same community and $q$ otherwise, both error rates and sparsity requirement may be improved \citep{ahn2018hypergraph}. However, such results highly rely on the restrictive homogeneous hSBM, and it remains unclear whether they can be extended to heterogeneous hSBM or the even more general HEM. 
		
	
	
	\section{Numerical experiments}
	
	We evaluate the performance of the proposed HEM method on a variety of synthetic and real-life non-uniform hypergraph network data. We compare its performance with some existing non-uniform hypergraph community detection methods in literature, including Tensor-SCORE \citep{dongxia2019}, spectral hypergraph partitioning (SHP; \cite{Ghoshdastidar_2017}), and weighted projection to graph method (WPTG; \cite{kumar2018hypergraph, GandD2015ICML}). Tensor-SCORE is designed for uniform hypergraph community detection and can be extended to non-uniform hypergraph by representing non-uniform hypergraph as a collection of uniform hypergraphs or by adding multiple null vertices to convert the non-uniform hypergraph to a uniform one as in \citet{ouvrard2017adjacency}. We abbreviate the former extension as TS-1, while the latter extension as TS-2. SHP converts a non-uniform hypergraph to an incident matrix, and then maximizes the hypergraph associativity or minimizes the normalized hypergraph cut. WPTG represents the non-uniform hypergraph by a weighted adjacency matrix, and then standard graph community detection methods such as the spectral clustering, SCORE \citep{Jin2015} and modularity maximization algorithms can be employed.
	
	Both HEM and Tensor-SCORE involve some tuning parameters, which can be optimally determined by some data-adaptive selection criteria, including the stability criteria \citep{junhui2010} or the network cross validation \citep{zhuji2020}. Yet such data-adaptive selection schemes can be computationally expensive. Alternatively, in our numerical experiments, we follow the treatment in \citet{dongxia2019} for the tuning parameters in Tensor-SCORE, and set $\lambda_n = 10^{-4}/n$ for HEM to prevent $J(\bm{\alpha})$ from vanishing too fast. We scale $\lambda_n$ properly in order to ensure more information can be learned from the network likelihood at the early iterations in the computing algorithm. The numerical performance of all the methods is assessed by the average scaled Hamming error in (\ref{hamming}).

	\subsection{Synthetic networks}
	
	We consider two scenarios of synthetic networks for numerical comparison.
	
	\textbf{Scenario 1}: The non-uniform hypergraph networks are generated from the HEM model in (\ref{logit}) and (\ref{LFM_Tensor}). We vary the number of vertices $n\in \{300, 400, 500\}$, the sparsity factor $s_n \in \{0.4, 0.2, 0.1, 0.05, 0.025\}$, and set the range of all the hypergraphs to be $m=3$ with $r = K =2$. First, we generate the community centers $C^* \in \mathbb{R}^{(K+1) \times r}$ with $C^*_k \sim N_r(\bm{0}_r, I_r)$ for $k \in [K]$ and the last row $C_{K+1} = r^{-1/2} \bm{1}_r$, where $I_r$ is the $r$-dimensional identity matrix. Next, the vertex community memberships $\psi^*_i$, $i\in [n]$, are generated from the multinomial distribution with parameters $\frac{1}{K}\bm{1}_K$ indicating that the communities are approximately of equal sizes. The community membership matrix $Z^* \in \{0, 1\}^{(n+1)\times (K+1)}$ can be constructed accordingly. After that, the hypergraph embedding matrix is generated as $\bm{\alpha}^* = Z^* C^* + \bm{\varepsilon}$. Herein, $\bm{\varepsilon}$ is a noise matrix with $\bm{\varepsilon}_{ij} \overset{i.i.d.}{\sim} N(0, 0.5^2)$ if $i \in [n]$ and $\bm{\varepsilon}_{ij} = 0$ if $i = n+1$. Finally, $\mathcal{P}^*$ can be computed based on $\bm{\alpha}^*$ . For $i_1, i_2, i_3 $ such that $\delta_{i_1i_2i_3}^{n+1, ord} =0$, the hyperedge among vertices $\{\{i_1,i_2,i_3\}\}\setminus\{n+1\}$ is generated with probability $p_{i_1i_2i_3}^*$ independently.
	
	\textbf{Scenario 2: } The hypergraph generation process is the same as in Scenario 1, except that community sizes can be unbalanced. Specifically, we fix $n = 300$, $m = 3$,  $s_n = 0.1$ and $r = K = 2$, and vary the parameters of the multinomial distribution $(\kappa_1, \kappa_2)$ such that $\kappa_1+\kappa_2 = 1$ and $\kappa_1/\kappa_2 \in \{0.1, 0.3, 0.5, 0.7, 0.9\}$. Clearly, as $\kappa_1/\kappa_2$ gets smaller, the communities become more unbalanced. 
	
	For both scenarios, the averaged scaled Hamming errors and their corresponding standard errors of various community detection methods over 50 independent replications are reported in Tables 1 and 2. 
	
	\begin{table}[!htbp]
		\begin{center}
			\caption{The averaged community detection errors and its standard errors over 50 independent replications of different methods on the non-uniform hypergraph networks generated from Scenario 1.}
			\begin{tabular}{p{0.5cm}|p{0.9cm}|p{2.5cm}p{2.4cm}p{2.4cm}p{2.4cm}p{2.4cm}}
				\hline
				$n$ & $s_n$ & HEM & TS-1 & TS-2 & SHP & WPTG\\			
				\hline
					\multirow{5}*{300} & 0.4 & \textbf{0.1052}(0.0152) & 0.1623(0.0200) & 0.1556(0.0170) & 0.2781(0.0270) & 0.2901(0.0200)\\
				& 0.2 & \textbf{0.0968}(0.0134) & 0.1788(0.0206) & 0.1561(0.0172) & 0.2823(0.0210) & 0.2874(0.0202)\\
				& 0.1 & \textbf{0.1026}(0.0149) & 0.1777(0.0197) & 0.1587(0.0172) & 0.2855(0.0206) & 0.2925(0.0200)\\
				& 0.05 & \textbf{0.1161}(0.0173) & 0.1891(0.0216) &0.1627(0.0184) & 0.2906(0.0202) & 0.2969(0.0192)\\
				& 0.025 & \textbf{0.1505}(0.0229) & 0.1797(0.0208) & 0.1659(0.0189) & 0.3018(0.0199) & 0.3077(0.0197)\\
				\hline
				\multirow{5}*{400} & 0.4 & \textbf{0.1094}(0.0159) & 0.1592(0.0200) & 0.1556(0.0167) & 0.2814(0.0204) & 0.2968(0.0197)\\
				& 0.2 & \textbf{0.1003}(0.0145) & 0.1780(0.0209) & 0.1566(0.0168) & 0.2825(0.0204) & 0.2976(0.0197)\\
				& 0.1 & \textbf{0.1120}(0.0155) & 0.1811(0.0196) &0.1557(0.0167) & 0.2868(0.0201) & 0.2987(0.0194)\\
				& 0.05 & \textbf{0.1251}(0.0184) & 0.1802(0.0200) & 0.1615(0.0175) &0.2918(0.0206) & 0.2996(0.0198) \\
				& 0.025 & \textbf{0.1342}(0.0213) & 0.1874(0.0215) & 0.1736(0.0197) & 0.2964(0.0201) & 0.3051(0.0196)\\
				\hline
					\multirow{5}*{500} & 0.4 & \textbf{0.1038}(0.0142) & 0.1407(0.0173)  & 0.1483(0.0155)  &  0.2768(0.0205) & 0.2848(0.0198)\\
				& 0.2 & \textbf{0.0927}(0.0125) & 0.1632(0.0195) & 0.1524(0.0163) & 0.2783(0.0205) & 0.2834(0.0197)\\
				& 0.1 & \textbf{0.0998}(0.0138) & 0.1688(0.0198) & 0.1524(0.0162) & 0.2827(0.0209) & 0.2873(0.0203)\\
				& 0.05 & \textbf{0.1112}(0.0153) & 0.1649(0.0182) & 0.1542(0.0167)& 0.2848(0.0208) & 0.2904(0.0203)\\
				& 0.025 & \textbf{0.1216}(0.0185) & 0.1683(0.0192) & 0.1561(0.0176) & 0.2916(0.0203)  & 0.2931(0.0198)\\
				\hline				
			
			\end{tabular}
		\end{center}
	\end{table}

\begin{table}[!htbp]
	\begin{center}
		\caption{The averaged community detection errors and their corresponding standard errors over 50 independent replications of different methods on the non-uniform hypergraph networks generated from Scenario 2.}
		\begin{tabular}{p{0.9cm}|p{2.5cm}p{2.4cm}p{2.4cm}p{2.4cm}p{2.4cm}}
			\hline
			$\kappa_1/\kappa_2$ & HEM & TS-1 & TS-2 & SHP & WPTG\\			
			\hline
		    0.1 & 0.2364(0.0233) & 0.3930(0.0179)  & \textbf{0.2236}(0.0246) & 0.3467(0.0184) & 0.3050(0.0225)\\
	        0.3 & \textbf{0.1169}(0.0166) & 0.2091(0.0259) & 0.1639(0.0216) & 0.3207(0.0211) & 0.3066(0.0218)\\
			0.5 & \textbf{0.1165}(0.0164) & 0.1834(0.0226) & 0.1541(0.0177) & 0.3107(0.0217) & 0.3021(0.0213)\\
			0.7 & \textbf{0.1012}(0.0147) & 0.1711(0.0218) & 0.1551(0.0172) & 0.3005(0.0216) & 0.3054(0.0209)\\
			0.9 & \textbf{0.1084}(0.0152) & 0.1725(0.0204) & 0.1595(0.0171) & 0.2927(0.0205) & 0.2907(0.0198)\\
			\hline
		\end{tabular}
	\end{center}
\end{table}

It is evident  that HEM yields the smallest community detection error among all the methods in both scenarios. The performance of HEM and Tensor-SCORE is much better than that of SHP and WPTG, mainly due to the fact that SHP and WPTG need to convert the hypergraph adjacency tensor to matrix and thus suffer from information loss. Further, the performance of TS-2 appears to be better than TS-1, suggesting that converting a non-uniform hypergraph to a uniform one by adding null vertices can be a better data processing approach than decomposing a non-uniform hypergraph into a collection of uniform hypergraphs of different range. It is also interesting to note that HEM is fairly robust to the hypergraph network sparsity and the community imbalance, whereas the performance of other competing methods can be substantially affected. 

To examine the network estimation accuracy, the averaged estimation errors of HEM, measured by $n^{-m/2} ||\widehat{\Theta} - \Theta^*||_F$, and their corresponding standard errors are reported in Table 3. Note that other competing methods solely focus on community detection, and thus do not produce estimate of $\Theta^*$.	Clearly, the estimation error of HEM becomes smaller as the number of vertices increases, the network becomes denser or the communities become more balanced. 
	
    \begin{table}[!htbp]
    	\begin{center}
    		\caption{The averaged estimation errors and their corresponding standard errors of HEM over 50 independent replications on the non-uniform hypergraph networks generated from Scenario 1 and 2.}
    		\begin{tabular}{p{1.9cm}|p{2.5cm}p{2.4cm}p{2.4cm}p{2.4cm}p{2.4cm}}
    			\hline
    			Scenario 1 &$s_n =$ 0.4 & 0.2 & 0.1 & 0.05 & 0.025\\			
    			\hline
    		   $n =300$ & 0.7165(0.1076) & 0.7462(0.1076)  & 0.7207(0.0977) & 0.8111(0.1174) & 0.9436(0.1252)\\
    			400 & 0.7072(0.1080) & 0.7192(0.1024) & 0.7368(0.1048) & 0.8182(0.1264) &  0.9472(0.1504)\\
    		    500 & 0.7301(0.1174) & 0.6965(0.1152) & 0.6820(0.1017) & 0.7468(0.1196) & 0.8027(0.1196) \\
    			\hline 
    			Scenario 2 & $\kappa_1/\kappa_2 = 0.1$ & 0.3 & 0.5 & 0.7 & 0.9\\
    			\hline
    			$n = 300$  & 1.2024(0.3076) & 0.8646(0.1824) & 0.8153(0.1585) & 0.7316(0.1382) & 0.7727(0.1221)\\  	
    			\hline		
    		\end{tabular}
    	\end{center}
    \end{table}
	
Finally, Figure 2 displays the first 15 eigenvalues of two randomly generated hypergraphs from Scenarios 1 and 2. It is clear that the first $2$ leading singular values  are substantially larger than the remaining singular values, confirming the choice of $K=2$ in the synthetic networks and the effectiveness of the eigen-gap approach.
	
	
	\begin{figure}[!htbp]
		\centering
		\includegraphics[width=6cm,height=4cm]{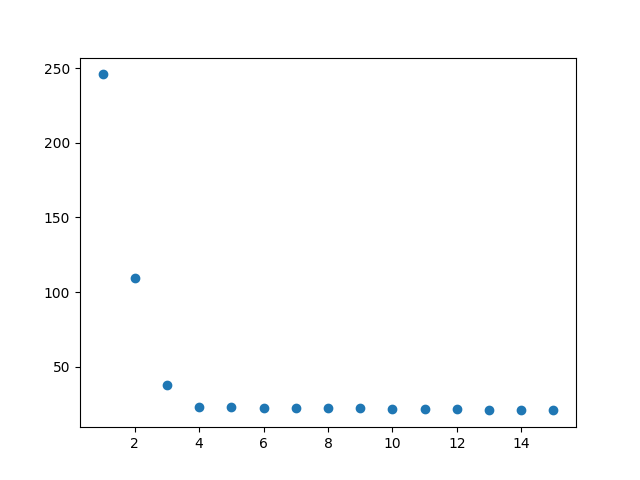}
		\includegraphics[width=6cm,height=4cm]{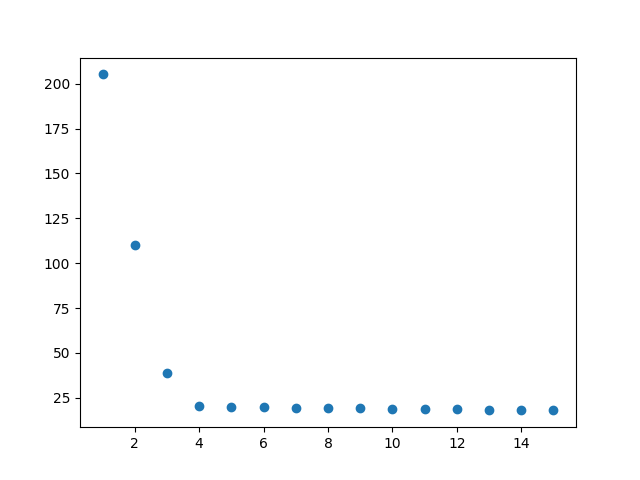}
		\caption{The first leading $15$ singular values of a hypergraph network randomly generated from Scenario 1 (left) and Scenario 2 (right).}
		\label{fig:eigengap}
	\end{figure}
	
	\subsection{Two real-life hypergraph networks}
	
	We now apply the proposed HEM method to analyze two real-life hypergraph networks, including the Medical Subject Headings (MeSH) hypergraph network \citep{dongxia2019} and the  cardiac Single Proton Emission Computed Tomography (SPECT) network \citep{Dua:2019}.
	
	The MeSH network is extracted from the MEDLINE database, which is available at https://www.nlm.nih.gov/bsd/medline.html. It consists of 318 MeSH terms of two diseases: Neoplasms (C04) and Nerve System Diseases (C10), which are represented as the vertices in the hypergraph network. The hyperedges are 10,472 papers published in 1960 where one or more of the above Mesh terms are annotated. After deleting the duplication hyperedges, there are a total of 1,375 hyperedges left, consisting of 297, 883, 174, 20 and 1 hyperedges of size 1, 2, 3, 4, and 5, respectively. We also remove those 21 hyperedges of size greater than 3 and all the hyperedges of size 1 due to their lack of information about latent community structure. After the pre-processing, we obtain a hypergraph of range 3 with 281 vertices and 1,057 hyperedges of size 2 or 3, where 180 vertices come from C04 and the other 101 vertices come from  C10.
	
	The SPECT network (https://archive.ics.uci.edu/ml/datasets/spect+heart) contains the SPECT images of 267 patients, where each image has been processed to 44 categorical features to discriminate abnormal patients from the normal ones.  To construct the hypergraph network, each patient is represented as a vertex, and for a possible value of each feature, we construct a hyperedge that contains all the patients sharing the particular feature value. Such a hypergraph construction method has been studied by \citet{Learingwithhypergraph2007} and \citet{Ghoshdastidar_2017}. After deleting the replication hyperedges, there are a total of 1,483 hyperedges with sizes verying from 1 to 36. We only consider those hyperedges of size ranging from 2 to 6, and for hyperedges with sizes greater than 3, we convert them to all possible 3-cliques, leading to multiple hyperedges with size 3. Replication hyperedges and isolated vertices are further deleted. After the pre-processing, we obtain a hypergraph with 264 vertices and 2,950 hyperedges, where 211 vertices come from the abnormal patients and the other 53 ones come from the normal patients.
	
	We perform different community detection methods on both hypergraph networks. As suggested in \cite{dongxia2019}, we set $K=6$ for the reg-HOOI algorithm in Tensor-SCORE, and thus we also set the embedding dimension $r=6$ in HEM for fair comparison. The network sparsity factor in HEM is estimated by hyperedge density; that is, $s_n=1057\times\big(\binom{281}{2} + \binom{281}{3}\big)^{-1}$ in MeSH network and $s_n=2950\times\big(\binom{264}{2} + \binom{264}{3}\big)^{-1}$ in SPECT network. The scaled Hamming errors of all the hypergraph community detection methods are reported in Table 4.
	
	\begin{table}[!htbp]
		\begin{center}
			\caption{Scaled Hamming errors of different hypergraph community detection methods on two real-life hypergraph networks.}
			\begin{tabular}{c|ccccc}
				\hline
				& HEM & TS-1 & TS-2 & SHP & WPTG\\
				\hline
				MeSH & \textbf{0.0427} & 0.0819 & 0.0925 & 0.0498 & 0.3630\\
				SPECT & \textbf{0.1970} & 0.3295 & 0.2159  & 0.3181 & 0.2652\\
				\hline
			\end{tabular}
		\end{center}
	\end{table}
	
\begin{figure}[!htbp]
		\includegraphics[width=8cm,height=6cm]{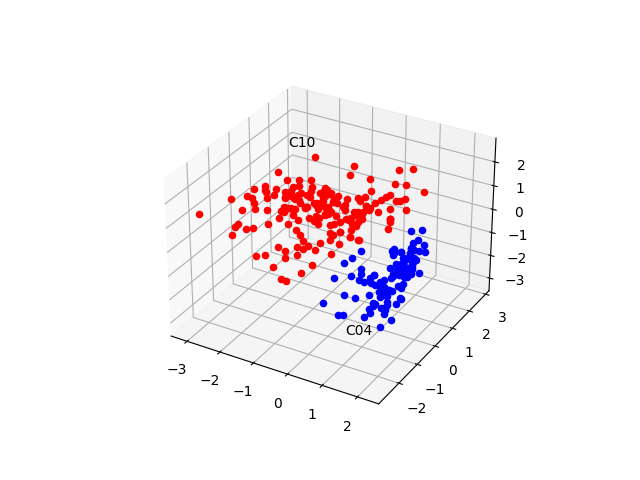}
		\includegraphics[width=8cm,height=6cm]{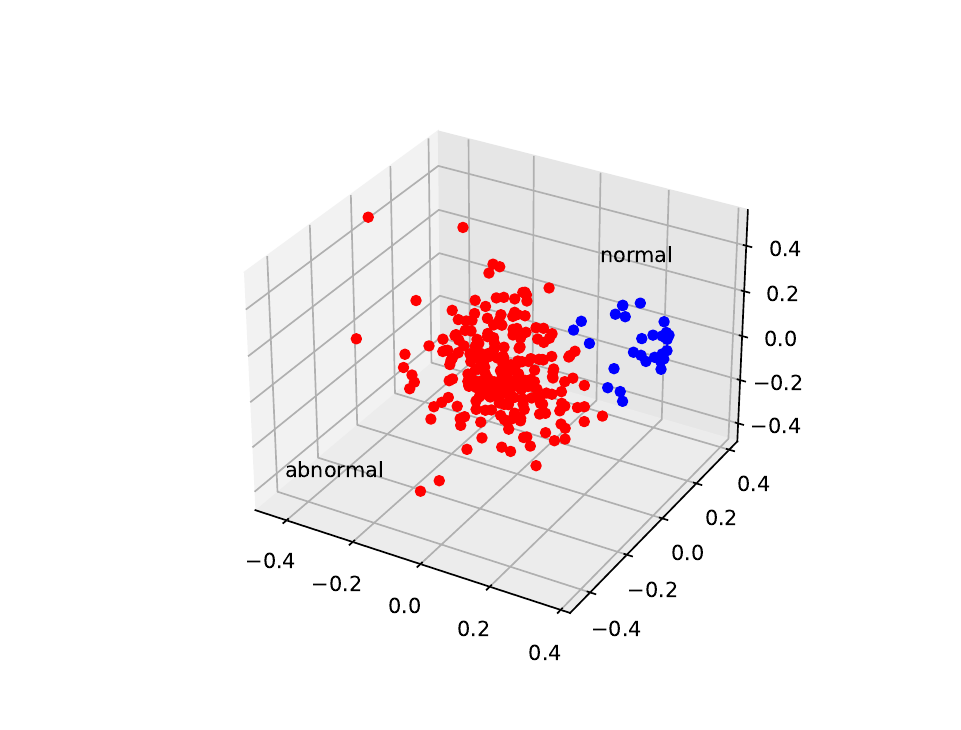}
		\caption{The detected communities in the embedding space of the MeSH hypergraph network (left) and the SPECT hypergraph network (right).}
		\label{fig:citation}
	\end{figure}

	It is evident that the scaled hamming errors of HEM are smaller than the other three competitors in both hypergraph networks, demonstrating its advantage in terms of community detection. We further visualize the estimated communities in both hypergraph networks in Figure \ref{fig:citation} by  multidimensional scaling, where the community structures detected by HEM are very clear in the embedding space.
	
	\section{Conclusion}
	
	This article proposes a novel community detection method on general hypergraph networks. The proposed method is built upon a tensor-based hypergraph embedding model, which consists of a network augmentation step and an network embedding step. The resultant negative log-likelihood function is equipped with a new regularization term to encourage community structures among the embedding vectors. The proposed method is supported by various numerical experiments and asymptotic consistency in terms of community detection. Particularly, the theoretical results can be established for very sparse hypergraph network with link probability of order $s_n\gg n^{1-m} \log n$. It is worth pointing out that the proposed community detection method can be extended to various scenarios, such as multi-layer hypergraph networks or communities with mixed memberships, which is under further investigation.

	\section*{Acknowledgment}
	This research is supported in part by HK RGC Grants GRF-11303918, GRF-11300919, and GRF-11304520. We thank the associate editor and three anonymous referees, whose constructive comments and suggestions have led to significant improvements of the article. We also thank Dr. Dong Xia for sharing the MeSH dataset. \\
	
	\begin{spacing}{1}
		
		\bibliography{ref}

\begin{thebibliography}{}

\bibitem[\protect\astroncite{Agarwal et~al.}{2005}]{agarwal2005beyond}
Agarwal, S., Lim, J., Zelnik-Manor, L., Perona, P., Kriegman, D., and Belongie,
  S. (2005).
\newblock Beyond pairwise clustering.
\newblock In {\em 2005 IEEE Computer Society Conference on Computer Vision and
  Pattern Recognition (CVPR'05)}, volume~2, pages 838--845. IEEE.

\bibitem[\protect\astroncite{Ahn et~al.}{2018}]{ahn2018hypergraph}
Ahn, K., Lee, K., and Suh, C. (2018).
\newblock Hypergraph spectral clustering in the weighted stochastic block
  model.
\newblock {\em IEEE Journal of Selected Topics in Signal Processing},
  12(5):959--974.

\bibitem[\protect\astroncite{Arous et~al.}{2019}]{Arous2019}
Arous, G.~B., Mei, S., Montanari, A., and Nica, M. (2019).
\newblock The landscape of the spiked tensor model.
\newblock {\em Communications on Pure and Applied Mathematics},
  72(11):2282--2330.

\bibitem[\protect\astroncite{Bahmanian and
  Sajna}{2015}]{bahmanian2015connection}
Bahmanian, M.~A. and Sajna, M. (2015).
\newblock Connection and separation in hypergraphs.
\newblock {\em Theory and Applications of Graphs}, 2(2):5.

\bibitem[\protect\astroncite{Chen and Yuan}{2006}]{chen2006detecting}
Chen, J. and Yuan, B. (2006).
\newblock Detecting functional modules in the yeast protein--protein
  interaction network.
\newblock {\em Bioinformatics}, 22(18):2283--2290.

\bibitem[\protect\astroncite{Chen and
  Lei}{2018}]{doi:10.1080/01621459.2016.1246365}
Chen, K. and Lei, J. (2018).
\newblock Network cross-validation for determining the number of communities in
  network data.
\newblock {\em Journal of the American Statistical Association},
  113(521):241--251.

\bibitem[\protect\astroncite{Chien et~al.}{2019}]{chien2019minimax}
Chien, I.~E., Lin, C.-Y., and Wang, I.-H. (2019).
\newblock On the minimax misclassification ratio of hypergraph community
  detection.
\newblock {\em IEEE Transactions on Information Theory}, 65(12):8095--8118.

\bibitem[\protect\astroncite{De~Lathauwer et~al.}{2000}]{HOSVD2000}
De~Lathauwer, L., De~Moor, B., and Vandewalle, J. (2000).
\newblock A multilinear singular value decomposition.
\newblock {\em SIAM Journal on Matrix Analysis and Applications},
  21(4):1253--1278.

\bibitem[\protect\astroncite{Dua and Graff}{2017}]{Dua:2019}
Dua, D. and Graff, C. (2017).
\newblock {UCI} machine learning repository.

\bibitem[\protect\astroncite{Gallo et~al.}{1993}]{directedhg1993}
Gallo, G., Longo, G., Pallottino, S., and Nguyen, S. (1993).
\newblock Directed hypergraphs and applications.
\newblock {\em Discrete Applied Mathematics}, 42(2):177 -- 201.

\bibitem[\protect\astroncite{Ghoshdastidar and Dukkipati}{2014}]{GandD2004NIPS}
Ghoshdastidar, D. and Dukkipati, A. (2014).
\newblock Consistency of spectral partitioning of uniform hypergraphs under
  planted partition model.
\newblock In Ghahramani, Z., Welling, M., Cortes, C., Lawrence, N.~D., and
  Weinberger, K.~Q., editors, {\em Advances in Neural Information Processing
  Systems 27}, pages 397--405. Curran Associates, Inc.

\bibitem[\protect\astroncite{Ghoshdastidar and
  Dukkipati}{2015a}]{GandD2015ICML}
Ghoshdastidar, D. and Dukkipati, A. (2015a).
\newblock A provable generalized tensor spectral method for uniform hypergraph
  partitioning.
\newblock In {\em Proceedings of the 32nd International Conference on
  International Conference on Machine Learning - Volume 37}, ICML’15, page
  400–409. JMLR.org.

\bibitem[\protect\astroncite{Ghoshdastidar and
  Dukkipati}{2015b}]{GandD2015AAAI}
Ghoshdastidar, D. and Dukkipati, A. (2015b).
\newblock Spectral clustering using multilinear svd: Analysis, approximations
  and applications.
\newblock In {\em Proceedings of the Twenty-Ninth AAAI Conference on Artificial
  Intelligence}, AAAI’15, page 2610–2616. AAAI Press.

\bibitem[\protect\astroncite{Ghoshdastidar and
  Dukkipati}{2017a}]{Ghoshdastidar_2017}
Ghoshdastidar, D. and Dukkipati, A. (2017a).
\newblock Consistency of spectral hypergraph partitioning under planted
  partition model.
\newblock {\em The Annals of Statistics}, 45(1):289–315.

\bibitem[\protect\astroncite{Ghoshdastidar and
  Dukkipati}{2017b}]{JMLR:v18:16-100}
Ghoshdastidar, D. and Dukkipati, A. (2017b).
\newblock Uniform hypergraph partitioning: Provable tensor methods and sampling
  techniques.
\newblock {\em Journal of Machine Learning Research}, 18(50):1--41.

\bibitem[\protect\astroncite{Hoff et~al.}{2002}]{hoff2002latent}
Hoff, P.~D., Raftery, A.~E., and Handcock, M.~S. (2002).
\newblock Latent space approaches to social network analysis.
\newblock {\em Journal of the American Statistical Association},
  97(460):1090--1098.

\bibitem[\protect\astroncite{Ji et~al.}{2016}]{ji2016coauthorship}
Ji, P., Jin, J., et~al. (2016).
\newblock Coauthorship and citation networks for statisticians.
\newblock {\em The Annals of Applied Statistics}, 10(4):1779--1812.

\bibitem[\protect\astroncite{Jin}{2015}]{Jin2015}
Jin, J. (2015).
\newblock Fast community detection by score.
\newblock {\em Ann. Statist.}, 43(1):57--89.

\bibitem[\protect\astroncite{Jing et~al.}{2021}]{jing2020community}
Jing, B.-Y., Li, T., Lyu, Z., and Xia, D. (2021).
\newblock Community detection on mixture multi-layer networks via regularized
  tensor decomposition.
\newblock {\em arXiv preprint arXiv:2002.04457}.

\bibitem[\protect\astroncite{Ke et~al.}{2021}]{dongxia2019}
Ke, Z.~T., Shi, F., and Xia, D. (2021).
\newblock Community detection for hypergraph networks via regularized tensor
  power iteration.
\newblock {\em arXiv preprint arXiv:1909.06503}.

\bibitem[\protect\astroncite{Kim et~al.}{2017}]{kim2017community}
Kim, C., Bandeira, A.~S., and Goemans, M.~X. (2017).
\newblock Community detection in hypergraphs, spiked tensor models, and
  sum-of-squares.
\newblock In {\em 2017 International Conference on Sampling Theory and
  Applications (SampTA)}, pages 124--128. IEEE.

\bibitem[\protect\astroncite{Kolda and Bader}{2009}]{SIAMreview}
Kolda, T.~G. and Bader, B.~W. (2009).
\newblock Tensor decompositions and applications.
\newblock {\em SIAM Review}, 51:455--500.

\bibitem[\protect\astroncite{Kova{\v{c}}evi{\'c} and
  Tan}{2018}]{kovavcevic2018codes}
Kova{\v{c}}evi{\'c}, M. and Tan, V.~Y. (2018).
\newblock Codes in the space of multisets—coding for permutation channels
  with impairments.
\newblock {\em IEEE Transactions on Information Theory}, 64(7):5156--5169.

\bibitem[\protect\astroncite{Kumar et~al.}{2021}]{kumar2018hypergraph}
Kumar, T., Vaidyanathan, S., Ananthapadmanabhan, H., Parthasarathy, S., and
  Ravindran, B. (2021).
\newblock Hypergraph clustering: a modularity maximization approach.
\newblock {\em arXiv preprint arXiv:1812.10869}.

\bibitem[\protect\astroncite{Lee et~al.}{2021}]{lee2020robust}
Lee, J., Kim, D., and Chung, H.~W. (2021).
\newblock Hypergraph clustering in the weighted stochastic block model via
  convex relaxation of truncated mle.
\newblock {\em arXiv preprint arXiv:2003.10038}.

\bibitem[\protect\astroncite{Lee et~al.}{2017}]{lee2017time}
Lee, S.~H., Magallanes, J.~M., and Porter, M.~A. (2017).
\newblock Time-dependent community structure in legislation cosponsorship
  networks in the congress of the republic of peru.
\newblock {\em Journal of Complex Networks}, 5(1):127--144.

\bibitem[\protect\astroncite{Lei et~al.}{2019}]{Leijing2020}
Lei, J., Chen, K., and Lynch, B. (2019).
\newblock {Consistent community detection in multi-layer network data}.
\newblock {\em Biometrika}, 107(1):61--73.

\bibitem[\protect\astroncite{Lei et~al.}{2015}]{lei2015consistency}
Lei, J., Rinaldo, A., et~al. (2015).
\newblock Consistency of spectral clustering in stochastic block models.
\newblock {\em Annals of Statistics}, 43(1):215--237.

\bibitem[\protect\astroncite{Li et~al.}{2020a}]{zhuji2020}
Li, T., Levina, E., and Zhu, J. (2020a).
\newblock {Rejoinder: ‘Network cross-validation by edge sampling’}.
\newblock {\em Biometrika}, 107(2):289--292.

\bibitem[\protect\astroncite{Li et~al.}{2020b}]{li2018convex}
Li, X., Chen, Y., and Xu, J. (2020b).
\newblock Convex relaxation methods for community detection.
\newblock {\em arXiv preprint arXiv:1810.00315}.

\bibitem[\protect\astroncite{Loyal and Chen}{2020}]{loyal2020statistical}
Loyal, J.~D. and Chen, Y. (2020).
\newblock Statistical network analysis: A review with applications to the
  coronavirus disease 2019 pandemic.
\newblock {\em International Statistical Review}, 88(2):419--440.

\bibitem[\protect\astroncite{Nepusz et~al.}{2012}]{nepusz2012detecting}
Nepusz, T., Yu, H., and Paccanaro, A. (2012).
\newblock Detecting overlapping protein complexes in protein-protein
  interaction networks.
\newblock {\em Nature methods}, 9(5):471.

\bibitem[\protect\astroncite{Newman et~al.}{2002}]{newman2002random}
Newman, M.~E., Watts, D.~J., and Strogatz, S.~H. (2002).
\newblock Random graph models of social networks.
\newblock {\em Proceedings of the National Academy of Sciences}, 99(suppl
  1):2566--2572.

\bibitem[\protect\astroncite{Ouvrard et~al.}{2021}]{ouvrard2017adjacency}
Ouvrard, X., Goff, J.-M.~L., and Marchand-Maillet, S. (2021).
\newblock Adjacency and tensor representation in general hypergraphs part 1:
  e-adjacency tensor uniformisation using homogeneous polynomials.
\newblock {\em arXiv preprint arXiv:1712.08189}.

\bibitem[\protect\astroncite{Pearson and Zhang}{2014}]{Pearson2014OnSH}
Pearson, K.~J. and Zhang, T. (2014).
\newblock On spectral hypergraph theory of the adjacency tensor.
\newblock {\em Graphs and Combinatorics}, 30:1233--1248.

\bibitem[\protect\astroncite{Pearson and Zhang}{2015}]{KellyJ.Pearson2015}
Pearson, K.~J. and Zhang, T. (2015).
\newblock The laplacian tensor of a multi-hypergraph.
\newblock {\em Discrete Mathematics}, 338(6):972 -- 982.

\bibitem[\protect\astroncite{Qi}{2005}]{QI20051302}
Qi, L. (2005).
\newblock Eigenvalues of a real supersymmetric tensor.
\newblock {\em Journal of Symbolic Computation}, 40(6):1302 -- 1324.

\bibitem[\protect\astroncite{Robeva}{2016}]{Robeva2016}
Robeva, E. (2016).
\newblock Orthogonal decomposition of symmetric tensors.
\newblock {\em SIAM Journal on Matrix Analysis and Applications},
  37(1):86--102.

\bibitem[\protect\astroncite{{Schölkopf}
  et~al.}{2007}]{Learingwithhypergraph2007}
{Schölkopf}, B., {Platt}, J., and {Hofmann}, T. (2007).
\newblock {\em Learning with Hypergraphs: Clustering, Classification, and
  Embedding}, pages 1601--1608.
\newblock MITP.

\bibitem[\protect\astroncite{Sengupta and Chen}{2018}]{sengupta2018block}
Sengupta, S. and Chen, Y. (2018).
\newblock A block model for node popularity in networks with community
  structure.
\newblock {\em Journal of the Royal Statistical Society: Series B (Statistical
  Methodology)}, 80(2):365--386.

\bibitem[\protect\astroncite{Sidiropoulos and
  Bro}{2000}]{sidiropoulos2000uniqueness}
Sidiropoulos, N.~D. and Bro, R. (2000).
\newblock On the uniqueness of multilinear decomposition of n-way arrays.
\newblock {\em Journal of Chemometrics: A Journal of the Chemometrics Society},
  14(3):229--239.

\bibitem[\protect\astroncite{Tang et~al.}{2020}]{tang2019individualized}
Tang, X., Xue, F., and Qu, A. (2020).
\newblock Individualized multi-directional variable selection.

\bibitem[\protect\astroncite{Tron and Vidal}{2007}]{tron2007benchmark}
Tron, R. and Vidal, R. (2007).
\newblock A benchmark for the comparison of 3-d motion segmentation algorithms.
\newblock In {\em 2007 IEEE conference on computer vision and pattern
  recognition}, pages 1--8. IEEE.

\bibitem[\protect\astroncite{Wang}{2010}]{junhui2010}
Wang, J. (2010).
\newblock {Consistent selection of the number of clusters via crossvalidation}.
\newblock {\em Biometrika}, 97(4):893--904.

\bibitem[\protect\astroncite{Yuan et~al.}{2021}]{yuan2018testing}
Yuan, M., Liu, R., Feng, Y., and Shang, Z. (2021).
\newblock Testing community structures for hypergraphs.
\newblock {\em arXiv preprint arXiv:1810.04617}.

\bibitem[\protect\astroncite{Zhang et~al.}{2021}]{zhang2021Directed}
Zhang, J., He, X., and Wang, J. (2021).
\newblock Directed community detection with network embedding.
\newblock {\em Journal of the American Statistical Association}, 0(0):1--11.

\bibitem[\protect\astroncite{Zhao et~al.}{2011}]{zhao2011community}
Zhao, Y., Levina, E., and Zhu, J. (2011).
\newblock Community extraction for social networks.
\newblock {\em Proceedings of the National Academy of Sciences},
  108(18):7321--7326.

\bibitem[\protect\astroncite{Zhao et~al.}{2012}]{zhao2012consistency}
Zhao, Y., Levina, E., Zhu, J., et~al. (2012).
\newblock Consistency of community detection in networks under degree-corrected
  stochastic block models.
\newblock {\em The Annals of Statistics}, 40(4):2266--2292.

\end{thebibliography}
		
	\end{spacing}
	
\end{document}